\documentclass[runningheads]{llncs}
\usepackage[fleqn]{amsmath}
\usepackage{graphicx}

\usepackage{times}
\usepackage{amssymb}
\usepackage{adjustbox}

\usepackage{subfigure}

\usepackage{multirow}
\usepackage{textcomp}
\usepackage{colortbl}
\usepackage{xcolor}
\usepackage{array,tabularx}
\usepackage{enumitem}
\usepackage{adjustbox}

\begin{document}
\title{RandomForestMLP: An Ensemble-Based Multi-Layer Perceptron Against Curse of Dimensionality}
\titlerunning{RandomForestMLP}
%
\author{Mohamed Mejri$^{1}$, Aymen Mejri$^{2}$}
\authorrunning{M. Mejri et al.}
%
\institute{School of Electrical and Computer Engineering, Georgia Institute of Technology, USA,$^{1}$\\
Image, Data and Signal department, Télécom Paris, Palaiseau, France, $^{2}$\\
\email{mohamed.mejri@gatech.edu$^{1}$\\aymen.mejri@telecom-paris.fr$^{2}$}}
\maketitle              
\begin{abstract}
We present a novel and practical deep learning pipeline termed RandomForestMLP. This core trainable classification engine consists of a convolutional neural network backbone followed by an ensemble-based multi-layer perceptrons core for the classification task. It is designed in the context of self and semi-supervised learning tasks to avoid overfitting while training on very small datasets. The paper details the architecture of the RandomForestMLP and present different strategies for neural network decision aggregation. Then, it assesses its robustness to overfitting when trained on realistic image datasets and compares its classification performance with existing regular classifiers.    

\keywords{Multi-layer perceptrons  \and Bagging techniques \and Image classification.}
\end{abstract}
\section{Introduction}

Modern computer vision systems have achieved outstanding performance on a variety of complex tasks such as image recognition, object detection, and semantic segmentation. Their success depends essentially on the availability of large annotated datasets. However, acquiring such a dataset is time-consuming and requires expensive storage capabilities. Real-world computer vision applications are often concerned with visual categories that are not present in standard benchmark datasets or with applications of a dynamic nature where visual categories or their appearance may change over time. Various semi- and self-supervised learning~\cite{zhai2019s4l} techniques have been used to automatically label the training set based on a small but reliable annotated dataset. However, as the dimensionality of the data points increases, especially when it comes to 2D images, automatic labeling techniques become inefficient due to the overfitting issue also known as the “curse of dimensionality.” In this paper, we propose a new classification method inspired by the random forest~\cite{freund1999short} algorithm and based on the fusion of several multi-layers perceptrons (MLPs) using features and data bagging techniques. This method will be assessed against benchmark datasets and then applied to examples of image classification.

\section{Related Work}

There is a body of research on models fusion and architecture aggregations. Bagging and specific training, data bootstrapping has been widely used as a
state-of-the-art ensemble-based classification technique.~\cite{HA200517} has designed an ensemble-based neural network architecture using training data bagging. Given a data set of N patterns, ~\cite{HA200517} builds L bootstrap samples by randomly sampling N patterns with replacement. Each bootstrap is fed forward to a single multilayered perceptron. The final model output is computed by averaging L outputs for a regression problem or by majority voting for a classification problem. Although bootstrapping aggregation is used to limit the impact of outliers on decision space construction and hence helps avoiding overfitting, feature-based bootstrapping introduced by ~\cite{freund1999short} and commonly used in the regular random forest~\cite{freund1999short} model reduces the effect of the curse of dimensionality, especially when it comes to the small training datasets . ~\cite{5521908} introduced the feature count measure that helps construct appropriate feature subspaces. His strategy helps increase the accuracy of ensemble-based decision trees when trained on UCI small datasets~\cite{Dua2019}. To the best of our knowledge, ensemble-based neural network architecture using the feature bagging technique does not exist. It extends the concept of feature bagging to neural networks.

\section{Context and Methods}

Several supervised learning approaches have been introduced and tested on various datasets. However, when it comes to automatically learning specific patterns from unlabeled datasets, several issues may occur, and the curse of dimensionality is one of them. Indeed, when training a relatively small dataset compared to its feature dimensionality, the classifier, especially a highly complex one, struggles to build a generalized decision space without overfitting. In image processing, one way to overcome this issue is to train a convolutional backbone to recognize the most important patterns inside the image, generate a feature map, and feed it forward to a simple multilayered perceptron to achieve classification. Our RandomForestMLP is based on the classical model of random forests~\cite{freund1999short} with decision trees. In contrast with the random forest~\cite{freund1999short}, which uses a specific number of classifiers and each one of them is trained using a random number of features (i.e., features bagging), the RandomForestMLP model uses all the subsets of N-1 features, where N refers to the total number of features.
\begin{center}
\begin{figure}
\includegraphics[width=0.8\textwidth]{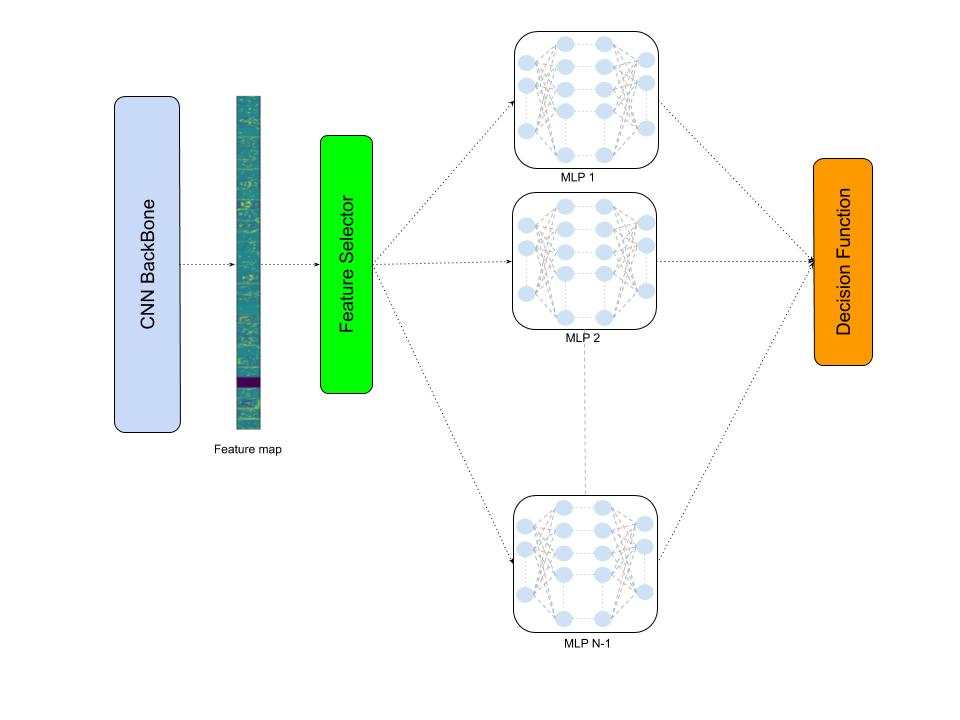}
\caption{An overview of the principal Components of the RandomForestMLP model} \label{fig1}
\end{figure}    
\end{center}

The figure \ref{fig1} shows the four principal components of the RandomForestMLP architecture: convolutional neural network (CNN) backbone, feature selector, MLP Forest, and decision function.
\subsection{CNN Backbone and Feature Selector algorithm}
A modified version of VGG16~\cite{simonyan2015deep} including batch normalization layers and using the parametric rectified linear~\cite{ramachandran2017searching} activation function instead of non-learnable rectified linear function was trained and its feature map was extracted and fed forward to the selection algorithm. It returned all the possible combinations of feature subsets. Each subset consists of N-1 features, where N refers to the total number of features in the feature map.                     
\subsection{MLP Forest and Decision Function}
It consists of N-1 multi-layer perceptron classifiers, each one is composed of one hidden layer followed by a fully connected layer. One MLP predictor is trained on the transformed feature map and returns a list of C probabilities where C is the number of classes. Those lists are fed forward to the decision function block to predict the final class member of each data point according to two different strategies. 
\subsubsection{Voting strategy}
The classification probabilities are filtered and binarized according to the following policy:\\
if ($max_{p_{c{j}}(i)}(p_{c{j}}(1),...,p_{c{j}}(C))>\frac{1}{2}$ then the classifier $c{j}$ decision is equal to\\ $argmax_{p_{c{j}}(i)}(p_{c{j}}(1),...,p_{c{j}}(C))$ otherwise it is rejected.\\
All the filtered class candidates are aggregated and we only chose the class with the major vote.
This voting strategy combined with the filtering process might provide more confident classification results and hence would be more robust to overfitting.
\subsubsection{Probabilistic Strategy}
This process is more subtle and relies on two different probabilistic approaches. The probability $P(y=i|x)$ to predict one class given its data point parameters is :\\
\begin{center}
$P(y=i|x) = \sum_{c_{j}\in C} P(y=i|c_{j},x_{j})P(c_{j})$
\end{center} where $C$ refers to the set of N-1 classifiers and $x_j$ to datapoint projected on their corresponding features.$P(y=i|c_{j},x_{j})$ corresponds to the $i-th$ perceptron value of the "Softmax" layer of the $j-th$ MLP classifier.
Since the classifiers are blind to the information value of each feature, we will consider that all the features are relevant and hence all the classifier's outcomes are equivalent. Therefore $P(c_{j})= \frac{1}{N_{C}}$ where $N_{C}$ refers to the dimensionality of the classifiers space.
However, the assumption of the classifier's blindness to the relevance of the features could be overcome by sorting the features according to their statistical information value, in other words by computing their variance values or formally by processing a principal component analysis decomposition of the original features. The covariance matrix of the training set X is $C=X^{T}X$. Its singular value decomposition is given by $C=P\Lambda.P^{T}$ where $\Lambda=Diag[\lambda_{1},...,\lambda_{N}]$ and $\lambda_{i}>\lambda_{j}$ if $i>j $\\ The transformed data is then $X^{'}=P\Lambda^{-1/2}X$. 
All the classifiers should take into consideration the relevance of the transformed features since they are sorted according to the eigenvalue ($\lambda_{i}$) of their covariance matrix $C$.
One way to achieve this operation is to penalize the classifiers that reject features with high eigenvalues according to the following classifiers probability distribution:
$P(c_{j})=\frac{\frac{1}{\lambda_{j}}}{\sum_{k=1}^{N}\frac{1}{\lambda_{k}}}$
\section{Experiments}
\subsection{Training datasets and materials}
The RandomForestMLP model without the Feature map extractor part was trained on several UCI datasets~\cite{Dua2019} such as (Iris plants dataset~\cite{Dua2019}, Wine recognition dataset~\cite{Dua2019}, Optical recognition of handwritten digits dataset~\cite{Dua2019}, etc). Each dataset was split according to their class members into K folds, and a cross-validation based assessment is performed: one fold is used for the training process, and K-1 are allocated to the validation process. K has to be large enough to induce a curse of dimensionality issue(i.e, the following condition must be satisfied $C^{N}<\frac{M}{K}$ where C, K, M, and N refers respectively to the number of different classes, the number of folds, the size of the training set and the dimensionality of the dataset).\\
Other realistic 2D image dataset (e.g, Cassava~\cite{mwebaze2019icassava} dataset, beans~\cite{beansdata} and Cars~\cite{KrauseStarkDengFei} dataset) were used to test the whole RandomForestMLP Pipeline classification performance.\\
All experiments were carried out on an IBM Power Systems AC922 with 256 GB of RAM and 4 NVIDIA V100 16 GB GPUs (using only a single GPU). Since on this platform, the GPU and the processors share a coherent memory space. We cannot guarantee that the previous settings are reproducible on Intel/AMD based machines as they may require smaller batches.
\subsection{Architectures experiments}
\begin{figure}
\begin{center}
\includegraphics[width=0.5\textwidth]{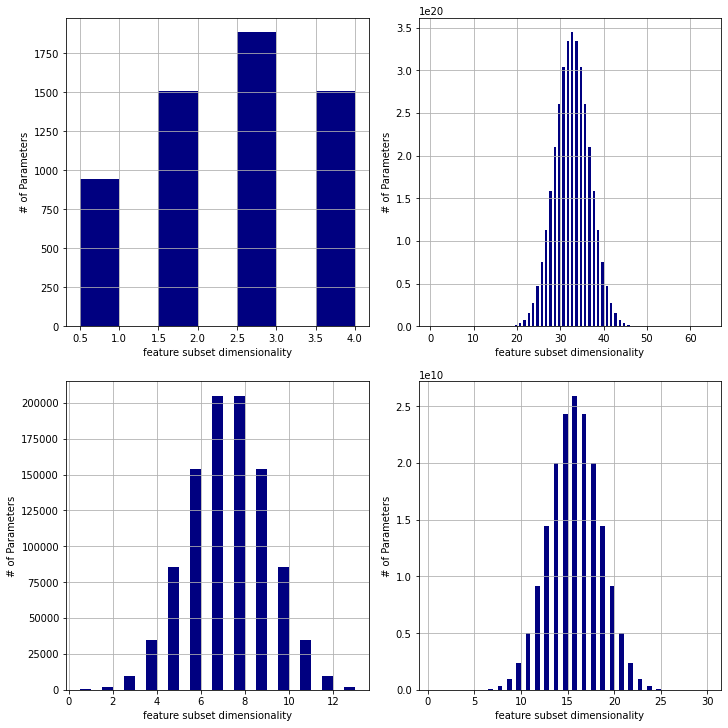}
\caption{The trainable parameters number of the MLPForest part depending on the feature subspace dimensionality for four datasets: (Upper Left) Iris dataset~\cite{Dua2019}, (Upper Right) digits dataset~\cite{Dua2019},(Lower Left) Wine dataset~\cite{Dua2019}, (Lower Right)breast cancer wisconsin dataset~\cite{Dua2019}} \label{fig2}
\end{center}
\end{figure}

Two different steps of experiments were achieved to assess the efficiency and the potential of our algorithm. they consist of respectively training the RandomForestMLP on small standardized datasets without the CNN backbone feature extractor then testing it on realistic image datasets using an enhanced version VGG16~\cite{simonyan2015deep} architecture to select and compress the relevant patterns of an image.
several multi-layer perceptrons algorithms were used to classify the standardized dataset. The exact number of classifiers depends only on the number of features subsets used. However, reducing the features space dimensionality would result in too many possible combinations of subsets and hence a huge number of classifiers. The figure\ref{fig2} shows the number of trainable parameters inside the MLPForest part and dimensionality of the feature sub-spaces of four different standardized datasets.


The number of parameters of the MLP Forest is proportional to the number of all possible feature subspaces with a given dimensionality(i.e, $(^{n}_{k})$) which become very large when $n>>1$ and $1<<k<<n$. This behavior is confirmed by figure \ref{fig2}.\\ 
Hence, we choose to keep only the feature space with very high dimensionality(i.e, $N_{f}$ = N-1 where $N_{f}$ and N refer respectively to the number of feature in the dataset and dimensionality of the feature subspace) to limit the computational expenses of the MLP Forest algorithm and to preserve feature information.\\
For computational purposes, each MLP Forest classifier consists of only one hidden layer, and each hidden layer is composed of 100 perceptrons.\\ 
The hyperparameter of the MLP predictors are tuned as follows: we chose an Adam~\cite{kingma2017adam} optimizer with an initial learning rate of $10^{-3}$, a learning rate scheduler is also used to avoid undesirable divergent behavior while increasing the number of epochs(the learning rate drops to $10^{-4}$ after 50 epochs).
Each MLP is trained on 200 batches of the same size for 100 epochs and an early stopping strategy was applied to avoid overfitting issues.\\
VGG16~\cite{simonyan2015deep} was used as a feature extractor backbone. It was first trained for 30 epochs on each dataset. Then, the weights of the CNN blocks were frozen and the final fully connected layer was removed.\\
\section{Results}
In this section, we will assess and compare the performance of our model with state of the art algorithms. We will demonstrate the capacity of the RandomForestMLP in reducing the overfitting effects when it comes to training with small datasets. The table below summarizes the weighted F1-score~\cite{f1score} of different classifiers trained on a small portion of the dataset (less than $20\%$) and tested on the rest.

\begin{table}[h]
\begin{center}
\caption{Classification performance(F1-Score~\cite{f1score}) of four different models after training on small subset of four standardized datasets} 
\label{tab:1}
\begin{tabular}{|c|l|l|l|l|l|l|}
\hline
\multirow{2}{*}{}               & \multicolumn{1}{c|}{\multirow{2}{*}{SVM$^{1}$\hspace{0.2cm}}} & \multirow{2}{*}{MLP$^{2}$\hspace{0.2cm}} & \multicolumn{1}{c|}{\multirow{2}{*}{RF$^{3}$\hspace{0.2cm}}} & \multicolumn{3}{c|}{Random Forest-MLP}                                                                                 \\ \cline{5-7} 
                                & \multicolumn{1}{c|}{}                                        &                                          & \multicolumn{1}{c|}{}                              & \multicolumn{1}{c|}{Majority Vote} & \multicolumn{1}{c|}{Equiprobability} & \multicolumn{1}{c|}{Weighted probability} \\ \hline
Iris Dataset                    & 64.1                                                         & 66.6                                     & 73.4                                               & 75.6                              & 76.7                                 & \bf79.2                                      \\ \hline
Digits Dataset                    & 47.3                                                         & 42.4                                     & 51.4                                               & 54.5                              & 52.3                                 & 54.1                                      \\ \hline
Wine Dataset                  & 45.1                                                         & 48.5                                     & 57.4                                               & 62.4                              & 63.1                                 & \bf65.8                                      \\ \hline
breast cancer Dataset & 53.7                                                         & 56.1                                     & 59.2                                               & 64.9                              & 66.1                                 & \bf68.2                                      \\ \hline
\end{tabular}
\end{center}
\footnotesize{$^1$ Support Vector Machines~\cite{Cortes1995SupportvectorN}, $^2$ Multi Layers Perceptron, $^3$ Random Forest~\cite{freund1999short}}
\end{table}
One major conclusion for the results given by the table \ref{tab:1} is that the Random-Forest-MLP model outperforms all the state of the art classifiers including random forest~\cite{freund1999short} with decision trees. However, the Weighted Probability strategy used in the decision function was effective and enhanced the F1-Score~\cite{f1score} of all the classification task by a minimum of $4\%$ compared to Major voting strategy and equiprobable feature space method except for Digits dataset where Majority vote acheives better classification performance.
Unlike Standardized datasets which are relatively easy to learn, 2D realistic images classification task is more complicated and will be a tougher assessment of our model. The table below compares the performance of a modified Version of VGG16~\cite{simonyan2015deep} architecture with our model using different decision function methods.
\\
\begin{table}[h]
\begin{center}
\caption{Classification performance (F1 score~\cite{f1score}) of four different models after being trained on datasets with small five-image subsets} 
\label{tab:2}
\begin{tabular}{|l|l|l|l|l|}
\hline
\multirow{2}{*}{}    & \multirow{2}{*}{VGG16~\cite{simonyan2015deep}} & \multicolumn{3}{c|}{RandomForestMLP(VGG16~\cite{simonyan2015deep} Backbone)}                                                                                 \\ \cline{3-5} 
                     &                                          & \multicolumn{1}{c|}{Majority Vote} & \multicolumn{1}{c|}{Equiprobability} & \multicolumn{1}{c|}{Weighted probability} \\ \hline
Flowers~\cite{tfflowers}              & 60.9                                     & 63.2                              & 63.4                                 & \bf64.5                                      \\ \hline
cars196~\cite{KrauseStarkDengFei}             & 35.2                                     & 43.1                              & 43.5                                 & \bf44.2                                      \\ \hline
Cassava~\cite{mwebaze2019icassava}              & 50.8                                     & 55.5                              & \bf55.6                                 & 53.7                                      \\ \hline
Beans~\cite{beansdata}                & 39.2                                     & 40.3                              & 51.2                                 & \bf59.6                                      \\ \hline
colorectal histology~\cite{kather2016multi} & 68.7                                     & \bf81.09                             & 81.08                                & 80.7                                      \\ \hline
\end{tabular}
\end{center}
\end{table}

\begin{figure}
\begin{center}
\includegraphics[width=0.5\textwidth]{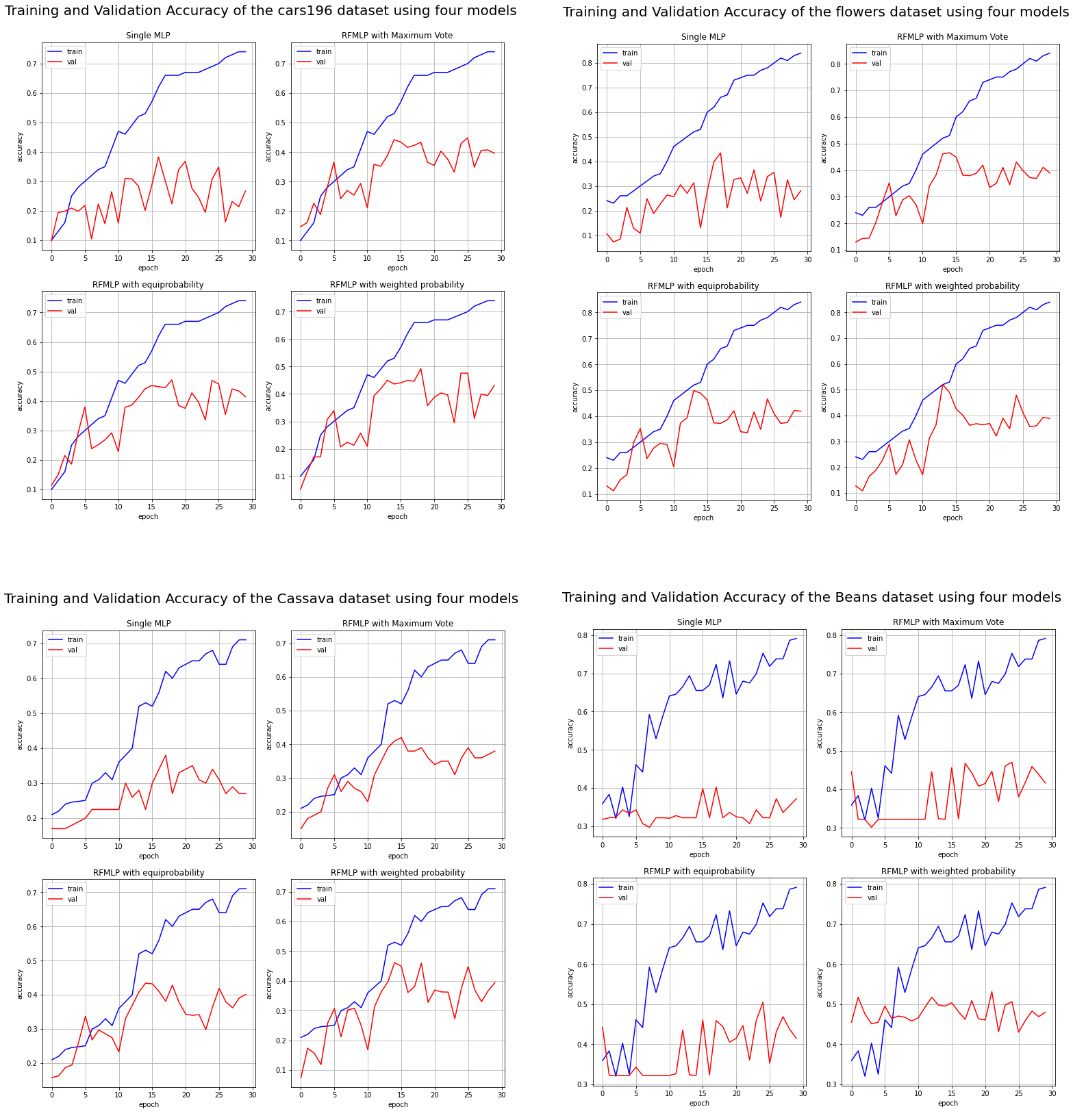}
\caption{ The variation of the training and the validation accuracy over 30 epochs of four different model after performing four different images classification tasks} \label{fig3}
\end{center}
\end{figure}

As for the standard dataset classification assessment, one major conclusion could be derived from the table \ref{tab:2}. Indeed, our model clearly outperforms the modified version of the VGG16~\cite{simonyan2015deep} on all the classification tasks. However, the choice of the decision function doesn't affect significantly the classification performance of the RandomForestMLP algorithm. The slight variance in F1-Score~\cite{f1score} between the three strategies is due to some features with high variance but less effective that misleads the classification balance strategy. Although the F1-scores~\cite{f1score} of the RandomForestMLP model shows its outstanding performance compared to regular CNN models when it comes to train on small portion of the dataset, it doesn't prove its robustness to overfitting. \\The four images in the figure~\ref{fig3} shows the variation of the validation and the training accuracy over 30 iterations using different models and image classification datasets. 
\\
From the figures we observe that the variation in validation accuracy of the regular VGG16~\cite{simonyan2015deep} does not follow the increase in training accuracy, which illustrates the overfitting issue. This behavior tends to reduce when the RandomForestMLP model is trained with its different decision function strategies


\newpage
\section{Conclusion}
In this paper, we have proved that an ensemble based neural network architecture with any decision function could outperform all the classification model build so far especially when training on small datasets. Indeed, while all regular classification models are affected by overfitting issues due to the curse of dimensionality in case of shortage of training data points, our model is more robust to this issue. RandomForestMLP usage could be extended to other realistic and more sophisticated tasks such as fine grained image segmentation and self-supervised 3D images classification.

%
%
%
 \bibliographystyle{splncs04}
%

\bibliography{ref}
\end{document}